\documentclass[conference]{IEEEtran}
\IEEEoverridecommandlockouts
\usepackage{cite}
\usepackage{amsmath,amssymb,amsfonts}
\usepackage{algorithmic}
\usepackage{mathtools}
\usepackage{graphicx}
\usepackage{textcomp}
\usepackage{xcolor}
\usepackage{booktabs}
\def\BibTeX{{\rm B\kern-.05em{\sc i\kern-.025em b}\kern-.08em
T\kern-.1667em\lower.7ex\hbox{E}\kern-.125emX}}
\usepackage[automake, nowarn]{glossaries}
\makeglossaries
\newacronym{ml}{ML}{machine learning}
\newacronym{fes}{FES}{Institute for Research and Development of Sports Equipment}
\newacronym{labp}{LaBP}{Laboratory for Biosignal Processing}

\newacronym{tn}{TN}{true negative}
\newacronym{fn}{FN}{false negative}
\newacronym{tp}{TP}{true positive}
\newacronym{fp}{FP}{false positive}
\newacronym{pla}{PLA}{piecewise linear approximation}

\newacronym[plural=CNNs, longplural=convolutional neural networks]{cnn}{CNN}{convolutional neural network}
\newacronym[plural=RNNs, longplural=recurrent neural networks]{rnn}{RNN}{recurrent neural network}
\newacronym[plural=GRUs, longplural=gated recurrent units]{gru}{GRU}{gated recurrent unit}
\newacronym[plural=BGRUs, longplural=bidirectional gated recurrent units]{bgru}{BGRU}{bidirectional gated recurrent unit}

\newacronym{cv}{CV}{cross validation}

\newacronym{relu}{ReLU}{rectified linear unit}

\newacronym[plural=TCNs, longplural=temporal convolutional networks]{tcn}{TCN}{temporal convolutional network}

\glsdisablehyper
\usepackage[caption=false]{subfig}
\begin{document}

\title{Using deep neural networks to detect non-analytically defined expert event labels \\ in canoe sprint force sensor signals}

\author{\IEEEauthorblockN{Sarah Rockstroh}
\IEEEauthorblockA{\textit{Laboratory for Biosignal Processing} \\
\textit{Leipzig University of Applied Sciences}\\
Leipzig, Germany \\
s.rockstroh@posteo.de}
\and
\IEEEauthorblockN{Patrick Frenzel}
\IEEEauthorblockA{\textit{Laboratory for Biosignal Processing} \\
\textit{Leipzig University of Applied Sciences}\\
Leipzig, Germany \\
patrick.frenzel@htwk-leipzig.de}
\and
\IEEEauthorblockN{Daniel Matthes}
\IEEEauthorblockA{\textit{Laboratory for Biosignal Processing} \\
\textit{Leipzig University of Applied Sciences}\\
Leipzig, Germany \\
daniel.matthes@htwk-leipzig.de}
\and
\IEEEauthorblockN{Kay Schubert}
\IEEEauthorblockA{\textit{Institute for Research and Development}\\ \textit{of Sports Equipment (FES)}\\
Berlin, Germany \\
kschubert@fes-sport.de}
\and
\IEEEauthorblockN{David Wollburg}
\IEEEauthorblockA{\textit{Institute for Research and Development}\\ \textit{of Sports Equipment (FES)}\\
Berlin, Germany \\
dwollburg@fes-sport.de}
\and
\IEEEauthorblockN{Mirco Fuchs*\thanks{*Corresponding author.}}
\IEEEauthorblockA{\textit{Laboratory for Biosignal Processing} \\
\textit{Leipzig University of Applied Sciences}\\
Leipzig, Germany \\
mirco.fuchs@htwk-leipzig.de}
}

\maketitle

\begin{abstract}
Assessing an athlete's performance in canoe sprint is often established by measuring a variety of kinematic parameters during training sessions.
Many of these parameters are related to single or multiple paddle stroke cycles. Determining on- and offset of these cycles in force sensor signals is usually not straightforward and requires human interaction. 
This paper explores \glspl{cnn} and \glspl{rnn} in terms of their ability to automatically predict these events. In addition, our work proposes an extension to the recently published SoftED metric for event detection in order to properly assess the model performance on time windows. In our results, an \gls{rnn} based on \glspl{bgru} turned out to be the most suitable model for paddle stroke detection.
\end{abstract}

\begin{IEEEkeywords}
canoe sprint, event detection, time series data, deep neural networks, machine learning
\end{IEEEkeywords}

\section{Introduction}\label{introduction}

In competitive canoe sprint the force exerted on the paddle is recorded and analyzed for the quantification~\cite{Gomes2015, Kong2020}, logging~\cite{Bonaiuto2022} and optimization~\cite{Michael2012}
of the technique of athletes. The corresponding force signals allow to derive parameters about the individual performance and potential for further performance improvements.

\begin{figure}[tb]
	\centering
	\includegraphics[width=0.9\linewidth]{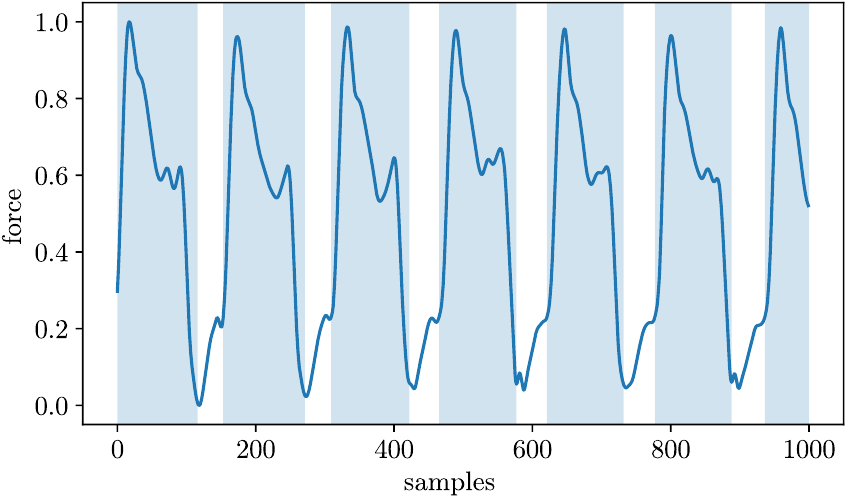}
	\caption{Blue line represents normalized paddle force signal, light blue shaded areas correspond to paddle cycles labeled by experts at FES. The signal shapes differ between athletes and boat/paddle type (canoe or kayak). Details in text.}
	\label{fig:force}
\end{figure}

Both canoe and kayak sprint obey a cyclic characteristic. Therefore, the analysis of an athlete's performance is based on segmenting time signals conducted during training runs into short sections, each corresponding to a single paddle stroke cycle.
Segmentation into individual paddle stroke cycles allows for numerous subsequent tasks as, e.\,g., assessing the curve shape throughout the paddle stroke~\cite{Baker1998, Bonaiuto2022, Gomes2015}, deducing additional stroke parameters ~\cite{Baker1998, McKenzie2019, Bonaiuto2022} and analyzing synchronicity between athletes~\cite{Gomes2015, Kong2020}.

The common analysis for German athletes conducted by sport scientists and service engineers at the \gls{fes} aims to detect the time span within each cycle in which the athlete attains a propulsive effect on the boat. The detection of the corresponding beginnings and ends of these segments cannot easily be achieved based on a well-defined analytic procedure like, e.\,g., analyzing thresholds and local extrema in the force signal and its derivatives. It rather requires the experts to identify these events in a mostly semi-automatic procedure using a customized algorithm. It comprises an initial very inaccurate automated prediction of all event locations in the whole signal and a subsequent iterative manual adjustment of various parameterized heuristic methods until the event locations are consistent with expert knowledge. Not only is this process very time consuming but also can it only be applied to the entire signal and not to shorter time windows such as only a few paddle cycles. An example of a force signal with expert labels is shown in Fig.~\ref{fig:force}. In this paper, we examine possibilities for realizing paddle stroke event detection using deep neural networks in a supervised learning setting. This approach allows for fully automatic paddle stroke detection and also paves the way for online detection.

A challenge in event detection is the measurement of the prediction performance. A suitable metric must account for the fact that any imperfectly detected event would count as a wrong prediction even if the error is on a $\pm 1$ sample scale. Therefore, a proper metric should consider the temporal distance between an event and the associated detection to weight the impact on the classification performance. A metric that accounts for the proximity of detections to their corresponding event labels and still allows the use of common classification performance measures such as accuracy, precision and recall is SoftED~\cite{Salles2023}. However, SoftED is less suited for windowed data since either labels or predictions close to beginning and end of the current window might have their corresponding counterpart either at the beginning of the subsequent window or at the end of the preceding window. This has a negative impact on the overall event detection performance. Hence, the contributions of this paper include

\begin{itemize}
\item the development and assessment of multiple CNN- and RNN-based neural network architectures for paddle stroke detection on force signals;
\item the extension of the SoftED metric for use on signal sections yielded by the sliding window method.
\end{itemize}

\begin{figure}[tb]
	\centering
	\includegraphics[width=0.95\linewidth]{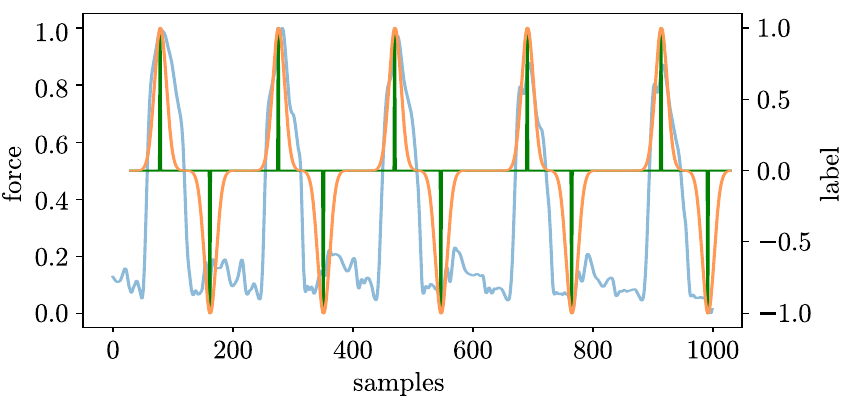}
	\caption{Blue: raw paddle force sensor signal; Green: encoded ternary labels, i.\,e. events. +1 denotes \textit{stroke beginning}, -1 denotes \textit{stroke ending}, and 0 denotes \textit{no event}. Orange: encoded labels after Gaussian smoothing (window length 100 samples, standard deviation 10~samples.)}
	\label{fig:labels}
\end{figure}

\section{Methods}\label{methods}

\subsection{Dataset}\label{dataset}

The force signals used to conduct the experiments were recorded by a measurement system developed at FES. It consists of one (for single-bladed canoe paddles) or two (for double-bladed kayak paddles) force sensors mounted at the paddle shaft and sampling at 200~Hz. 
The sensors are equipped with a radio module allowing for data transmission to an onboard computer.

The dataset used in this work consists of measurements from 109 canoe and kayak runs labeled by three experts at FES. Each run may include force signals from one to four athletes. We applied the following steps for preprocessing: (1) linear interpolation of transmission errors; (2) extraction of equally long time segments using sliding windows (window length $=1000$~samples, stride $=100$); and (3) min-max normalization to each window. 
The data was split into a train and a holdout set. 5-fold cross validation was used on the train set to identify the best performing model given a certain network architecture (i.\,e., CNN or RNN). These best performing models were then trained on the train set and compared during final evaluation on the holdout data. We applied a subject-aware split such that measurements from an athlete are only used for either training, validation \textit{or} test.

\subsection{Label Encoding}\label{label-encoding}

\begin{table}[b] 
	\centering 
	\caption{CNN + dense}
	\label{tab:CNN_plus_dense} 
	\begin{tabular}{llrl} 
		\toprule
		layer (type)                                            & output shape                & \# parameters 	& activation    \\ \midrule
		conv1d (Conv1D)                                          & (None, 1000, 64)            & 256       	& ReLU \\
		conv1d\_1 (Conv1D)                                       & (None, 1000, 64)            & 12352     	& ReLU \\ \midrule
		conv1d\_2 (Conv1D)                                       & (None, 1000, 128)           & 24704     	& ReLU \\ \midrule
		conv1d\_3 (Conv1D)                                       & (None, 1000, 256)           & 98560     	& ReLU \\ \midrule
		conv1d\_4 (Conv1D)                                       & (None, 1000, 512)           & 393728    	& ReLU \\
		conv1d\_5 (Conv1D)                                       & (None, 1000, 512)           & 786944    	& ReLU \\ \midrule
		flatten (Flatten)                                        & (None, 512000)              & 0         	& - \\
		dense (Dense)                                            & (None, 1000)                & 512001000 	& linear \\ \midrule
		\multicolumn{2}{l}{parameters: 513,317,544}       \\ \bottomrule
	\end{tabular} 
\end{table}
Our encoding procedure is shown in Fig.~\ref{fig:labels}.
The ternary event labels (stroke beginning, stroke ending, no event) were encoded into a 1D vector with the same length as the sample window with +1 denoting \textit{stroke beginning}, 0~denoting \textit{no event} and~-1 denoting \textit{stroke ending}.
Additionally, we adopt the approach of smoothing the discrete labels using a Gaussian filter that is common in the 2D domain~\cite{Tompson2014, Wei2016, Tremblay2018}.

\subsection{Convolutional Neural Networks}\label{convolutional-neural-networks}

The use of CNNs has proven effective in various domains~\cite{Lecun1998, Redmon2016, Yamashita2018, Kiranyaz2021}.
In this work, we investigate several CNN-based architectures in terms of their eligibility for paddle stroke detection in force sensor signals.
The general construction scheme to define a particular CNN model is based on the main concepts of the well-known VGG architecture~\cite{Simonyan2015}:
First, we use an increasing number of filters from input to output.
Second, multiple subsequent layers each with the same number of kernels form blocks.
Finally, a kernel of size three and same padding is used in each convolutional layer.
Four models have been derived using this scheme, they are presented below.

Table~\ref{tab:CNN_plus_dense} shows the first model, (\textit{CNN + dense}).
It consists of six convolutional layers, a flatten layer and is terminated by a dense layer.
Table~\ref{tab:CNN_c1} presents \textit{CNNc1}, a fully convolutional network. It only contains  convolutional layers. The single kernel in the terminating convolutional layer is used to adjust the output to the input shape. 

\textit{CNNc2} (Tab.~\ref{tab:CNN_c2}) and \textit{CNNc3} (Tab.~\ref{tab:CNN_c3}) are extensions of \textit{CNNc1} employing a higher number of layers per block and more filters per layer.
All further design principles remain unchanged from \textit{CNNc1}.
These deeper variants are intended to indicate whether increased model complexity might be beneficial for the quality of paddle stroke detection.

\begin{table}[tb] 
	\centering 
	\caption{CNNc1} 
	\label{tab:CNN_c1} 
	\begin{tabular}{llrl} 
		\toprule
		layer (type)                                            & output shape                & \# parameters   & activation \\ \midrule
		conv1d (Conv1D)                                          & (None, 1000, 64)            & 256       & ReLU \\
		conv1d\_1 (Conv1D)                                       & (None, 1000, 64)            & 12352     & ReLU \\ \midrule
		conv1d\_2 (Conv1D)                                       & (None, 1000, 128)           & 24704     & ReLU \\ \midrule
		conv1d\_3 (Conv1D)                                       & (None, 1000, 256)           & 98560     & ReLU \\ \midrule
		conv1d\_4 (Conv1D)                                       & (None, 1000, 512)           & 393728    & ReLU \\
		conv1d\_5 (Conv1D)                                       & (None, 1000, 512)           & 786944    & ReLU \\ \midrule
		conv1d\_6 (Conv1D)                                       & (None, 1000, 1)             & 513       & linear \\ \midrule
		\multicolumn{2}{l}{parameters: 1,317,057}         \\ \bottomrule
	\end{tabular} 
\end{table}

\begin{table}[tb]
	\centering 
	\caption{CNNc2}
	\label{tab:CNN_c2} 
	\begin{tabular}{llrl} 
		\toprule
		layer (type)                                            & output shape                & \# parameters  & activation  \\ \midrule
		conv1d (Conv1D)                                          & (None, 1000, 64)            & 256       & ReLU \\
		conv1d\_1 (Conv1D)                                       & (None, 1000, 64)            & 12352     & ReLU \\ \midrule
		conv1d\_2 (Conv1D)                                       & (None, 1000, 128)           & 24704     & ReLU \\
		conv1d\_3 (Conv1D)                                       & (None, 1000, 128)           & 49280     & ReLU \\ \midrule
		conv1d\_4 (Conv1D)                                       & (None, 1000, 256)           & 98560     & ReLU \\
		conv1d\_5 (Conv1D)                                       & (None, 1000, 256)           & 196864    & ReLU \\ \midrule
		conv1d\_6 (Conv1D)                                       & (None, 1000, 512)           & 393728    & ReLU \\
		conv1d\_7 (Conv1D)                                       & (None, 1000, 512)           & 786944    & ReLU \\ \midrule
		conv1d\_8 (Conv1D)                                       & (None, 1000, 1024)          & 1573888   & ReLU \\
		conv1d\_9 (Conv1D)                                       & (None, 1000, 1024)          & 3146752   & ReLU \\ \midrule
		conv1d\_10 (Conv1D)                                      & (None, 1000, 1)             & 1025      & linear \\ \midrule
		\multicolumn{2}{l}{parameters: 6,284,353}         \\ \bottomrule
	\end{tabular} 
\end{table}

\begin{table}[tb] 
	\centering 
	\caption{CNNc3}
	\label{tab:CNN_c3} 
	\begin{tabular}{llrl} 
		\toprule
		layer (type)                                            & output shape                & \# parameters   & activation \\ \midrule
		conv1d (Conv1D)                                          & (None, 1000, 64)            & 256       & ReLU \\
		conv1d\_1 (Conv1D)                                       & (None, 1000, 64)            & 12352     & ReLU \\
		conv1d\_2 (Conv1D)                                       & (None, 1000, 64)            & 12352     & ReLU \\ \midrule
		conv1d\_3 (Conv1D)                                       & (None, 1000, 128)           & 24704     & ReLU \\
		conv1d\_4 (Conv1D)                                       & (None, 1000, 128)           & 49280     & ReLU \\
		conv1d\_5 (Conv1D)                                       & (None, 1000, 128)           & 49280     & ReLU \\ \midrule
		conv1d\_6 (Conv1D)                                       & (None, 1000, 256)           & 98560     & ReLU \\
		conv1d\_7 (Conv1D)                                       & (None, 1000, 256)           & 196864    & ReLU \\
		conv1d\_8 (Conv1D)                                       & (None, 1000, 256)           & 196864    & ReLU \\ \midrule
		conv1d\_9 (Conv1D)                                       & (None, 1000, 512)           & 393728    & ReLU \\
		conv1d\_10 (Conv1D)                                      & (None, 1000, 512)           & 786944    & ReLU \\
		conv1d\_11 (Conv1D)                                      & (None, 1000, 512)           & 786944    & ReLU \\ \midrule
		conv1d\_12 (Conv1D)                                      & (None, 1000, 1024)          & 1573888   & ReLU \\
		conv1d\_13 (Conv1D)                                      & (None, 1000, 1024)          & 3146752   & ReLU \\
		conv1d\_14 (Conv1D)                                      & (None, 1000, 1024)          & 3146752   & ReLU \\ \midrule
		conv1d\_15 (Conv1D)                                      & (None, 1000, 1)             & 1025      & linear \\ \midrule
		\multicolumn{2}{l}{parameters: 10,476,545}        \\ \bottomrule
	\end{tabular} 
\end{table}

\subsection{Recurrent Neural Networks}\label{recurrent-neural-networks}

In contrast to CNNs, RNNs are particularly suited to process sequential input data and therefore also used in this work. We focus on \glspl{gru} which avoid exploding or vanishing gradients while decreasing hidden cell complexity compared to former RNNs~\cite{Goodfellow2016}. Table~\ref{tab:GRUc1} shows the first \gls{gru}, i.\,e. \textit{GRUc1}.
It consists of two unidirectional \glspl{gru} terminated by a dense layer.
Each \gls{gru} contains 64 units.

The position of a paddle stroke event in force sensor signals depends both on preceding and subsequent samples in the vicinity of an event. Therefore, we also employ a model based on \glspl{bgru}. 
Table~\ref{tab:BGRUc1} shows \textit{BGRUc1}, a bidirectional adaptation of \textit{GRUc1}.
Again, every \gls{bgru} contains 64 units.
However, the outputs of both \glspl{gru} per layer are concatenated.
Therefore, the size of the second output dimension is doubled (as per Tab.~\ref{tab:BGRUc1}).

Likewise to CNNs, we used two more complex architectures.
\textit{BGRUc2} (Tab.~\ref{tab:BGRUc2}) is made up of four \glspl{bgru}, each containing 64 units.
\textit{BGRUc3} (Tab.~\ref{tab:BGRUc3}) also contains four \glspl{bgru}; however, each \gls{bgru} includes 128 units.

\begin{table}[tb] 
	\centering 
	\caption{GRUc1}
	\label{tab:GRUc1} 
	\begin{tabular}{llrl} 
		\toprule 
		layer (type)                                        & output shape                & \# parameters  & activation  \\ \midrule
		gru (GRU)                                            & (None, 1000, 64)            & 12864         & tanh \\
		gru\_1 (GRU)                                         & (None, 1000, 64)            & 24960         & tanh \\ \midrule
		dense (Dense)                                        & (None, 1000, 1)             & 65            & linear \\ \midrule
		\multicolumn{2}{l}{parameters: 37,889}        \\ \bottomrule
	\end{tabular} 
\end{table}

\begin{table}[tb] 
	\centering 
	\caption{BGRUc1}
	\label{tab:BGRUc1} 
	\begin{tabular}{llrl} 
		\toprule 
		layer (type)                                        & output shape               & \# parameters  & activation  \\ \midrule
		bidirectional (BGRU)                           & (None, 1000, 128)          & 25728         & tanh \\
		bidirectional\_1 (BGRU)                        & (None, 1000, 128)          & 74496         & tanh \\ \midrule
		dense (Dense)                                        & (None, 1000, 1)            & 129           & linear \\ \midrule
		\multicolumn{2}{l}{parameters: 100,353}       \\ \bottomrule
	\end{tabular} 
\end{table}

\begin{table}[tb] 
	\centering 
	\caption{BGRUc2}
	\label{tab:BGRUc2} 
	\begin{tabular}{llrl} 
		\toprule 
		layer (type)                                        & output shape                & \# parameters  & activation  \\ \midrule
		bidirectional (BGRU)                           & (None, 1000, 128)           & 25728         & tanh \\
		bidirectional\_1 (BGRU)                        & (None, 1000, 128)           & 74496         & tanh \\
		bidirectional\_2 (BGRU)                        & (None, 1000, 128)           & 74496         & tanh \\
		bidirectional\_3 (BGRU)                        & (None, 1000, 128)           & 74496         & tanh \\ \midrule
		dense (Dense)                                        & (None, 1000, 1)             & 129           & linear \\ \midrule
		\multicolumn{2}{l}{parameters: 249,345}       \\ \bottomrule
	\end{tabular} 
\end{table}

\begin{table}[tb] 
	\centering 
	\caption{BGRUc3}
	\label{tab:BGRUc3} 
	\begin{tabular}{llrl} 
		\toprule 
		layer (type)                                        & output shape                & \# parameters  & activation  \\ \midrule
		bidirectional (BGRU)                           & (None, 1000, 256)           & 100608        & tanh \\
		bidirectional\_1 (BGRU)                        & (None, 1000, 256)           & 296448        & tanh \\
		bidirectional\_2 (BGRU)                        & (None, 1000, 256)           & 296448        & tanh \\
		bidirectional\_3 (BGRU)                        & (None, 1000, 256)           & 296448        & tanh \\ \midrule
		dense (Dense)                                        & (None, 1000, 1)             & 257           & linear \\ \midrule
		\multicolumn{2}{l}{parameters: 990,209}       \\ \bottomrule
	\end{tabular} 
\end{table}

\subsection{Postprocessing and event detection}\label{post-processing-and-event-detection}

\begin{figure}[tb]
	\centering
	\includegraphics[width=\linewidth]{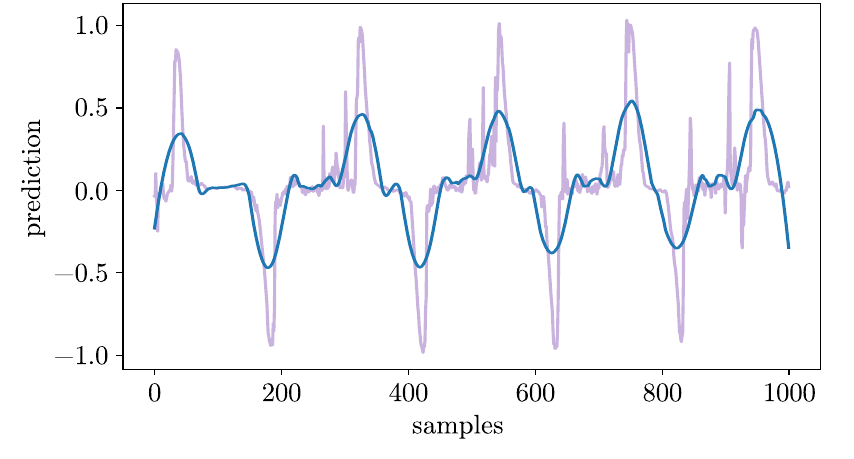}
	\caption{Filtering of raw neural network output. The raw prediction (purple) can be fairly noisy. Therefore, the model outputs are filtered using a second order Savitzky-Golay filter. The result is illustrated in blue.}
	\label{fig:prediction_filtering}
\end{figure}

Each sample in the predicted model outputs corresponds to a value between $-1$ and $+1$ (cf.~Sec~\ref{label-encoding}). 
Since the predicted signal can be fairly noisy (see Fig.~\ref{fig:prediction_filtering}), the raw model outputs are smoothed using a second order Savitzky-Golay low-pass filter before further postprocessing.

\begin{figure}[tb]
	\centering
	\includegraphics[width=\linewidth]{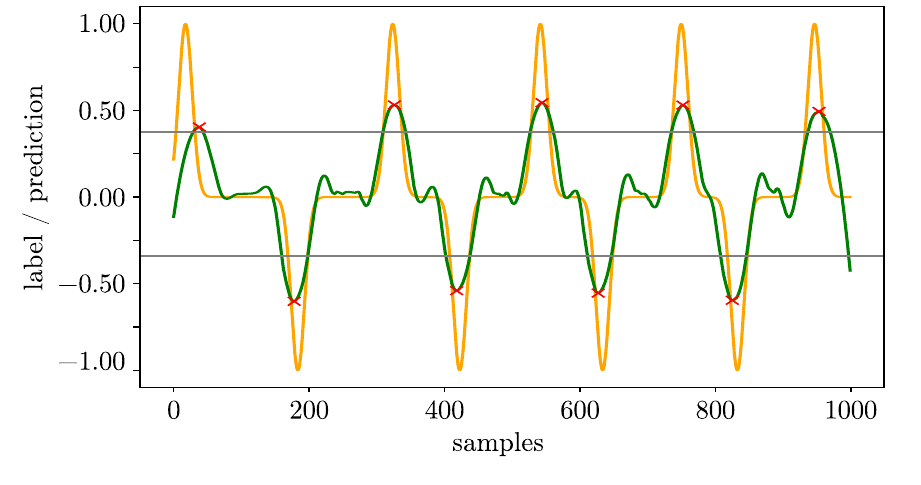}
	\caption{Event selection from filtered model output: (1) all local maxima above the 85\textsuperscript{th} percentile of positive values (gray line) are considered paddle stroke beginnings; (2) all local minima below the 15\textsuperscript{th} percentile of negative output values (lower gray line) are considered paddle stroke endings.}
	\label{fig:prediction_peaks}
\end{figure}

The events to be extracted correspond to the set of major minima and maxima in the predicted and smoothed model outputs.
Their detection is illustrated in Fig.~\ref{fig:prediction_peaks}.
First, an upper threshold is determined as the 85\textsuperscript{th} percentile of all positive output values.
All local maxima exceeding this threshold are considered paddle stroke onsets.
Second, a lower threshold is determined as the 15\textsuperscript{th} percentile of all negative output values.
All local minima below this threshold are considered paddle stroke endings.

To reduce the number of false positive events, detections close in time are combined into a single detection. 
This is realized by clustering. All detections in the vicinity of $\le$~5~samples to another detection are grouped and only the detection with the highest output value is retained while all others are dismissed.
If multiple detections share the same output value, their temporal average is used as the representative detection.

\subsection{Extending SoftED for sliding window method}\label{extending-softed-for-sliding-window-method}

The SoftED metric proposed in \cite{Salles2023} defines a membership function for each event in a time series. Predicted events temporally close to a labeled event are considered as members, i.\,e. they are associated with each other. Therefore, both the event predicted by a model and its associated event defined by a label have to be contained in the set of all predictions and labels used in the association process. If either the predicted event or the target event are not part of this set, they cannot be associated to each other. While this is usually justified when analyzing some signal in a holistic manner, it might not be the case if only parts of the signal, e.g., obtained using sliding windows are used. Consider the following scenario:
An event is detected right at the beginning of the window.
The corresponding event might be located even before the beginning of that window, i.\,e. in the ``previous'' window.

We extended SoftED to only evaluate detections within a valid range of the time window. In order to prevent another hard margin problem, we exclude not only detections in the margins of a window but also their corresponding events, no matter whether the event is located in the margin or valid window. Likewise, events in the margins of time windows are excluded from evaluation as well as their associated detections.
For our investigation, a margin width of 15~samples is used.

The procedure is as follows. Note that all symbols used are in accordance to~\cite{Salles2023}.
Let $l$ denote the time index, $t$ the samples and $h$ the width of the margin on each end of the window.
Then
\begin{equation} \label{eq:T_prime}
	T' = \{ t_l \mid h < l \le (|t|-h) \}	
\end{equation}

yields the reduced set $T'$ of time values that are considered for event detection.
This allows for the elimination of events/detections outside the valid range of the window:
\begin{align}
	E' &= \{e_j \mid t_{e_j} \in T' \land ( t_{d_i} \in T'\, \forall\, d_i \in D_{e_j} ) \} \label{eq:E_prime} \\
	D' &= \{d_i \mid t_{d_i} \in T' \land ( t_{e_j} \in T'\, \forall\, e_j \in E_{d_i} ) \} \label{eq:D_prime}
\end{align}

Equation~\refeq{eq:E_prime} yields the set of events $E'$ that are both located in the valid range $T'$ and only have assigned detections $d_i$ within the valid range $T'$.
Equivalently, Eq.~\refeq{eq:D_prime} yields the set of detections $D'$ located in the valid range $T'$ with all corresponding events $e_j$ within the valid range $T'$.

For consistency with traditional SoftED, the total number of time values $|t|$ must be reduced by the number of time values in the margins, i.\,e. $|t'| = |t| - 2h$. Next, events, detections and the number of time values from traditional SoftED can be replaced with our version, i.\,e. $E \coloneqq E'$, $D \coloneqq D'$ and $t \coloneqq t'$. Finally, $\text{TP}_S$, $\text{FP}_S$, $\text{FN}_S$ and $\text{TN}_S$ can be determined as per~\cite{Salles2023}.

\subsection{Experimental setup}\label{Experimental-setup-and-target-metrics}

We implemented the models presented in Sec.~\ref{convolutional-neural-networks} and~\ref{recurrent-neural-networks} using \textit{Keras} and \textit{Tensorflow}.
Training was performed as described in Sec.~\ref{dataset} using cluster nodes with 16\,GB RAM and 11\,GB GPU memory.

Evaluation is based on the soft versions of recall, precision and $F_1$ proposed in~\cite{Salles2023}. We derived $\text{TP}_S$, $\text{FP}_S$, $\text{FN}_S$ and $\text{TN}_S$ as described in Sec.~\ref{extending-softed-for-sliding-window-method}.
Recall, precision and $F_1$ are chosen as target metrics because of the unbalanced nature of paddle stroke detection problem:
While only few events may occur during a time window comprised to one or several paddle stroke cycles, the number samples corresponding to \textit{no event} is many times higher.
The tolerance $k$ for event detection (cf.~\cite{Salles2023}) is set to 15~samples.

Additionally, we consider the distribution of SoftED scores by analyzing their histograms.
This allows for the detection of possible strengths or weaknesses of the models, e.\,g. a high number of close detections.

\section{Results}\label{results}

Table~\ref{tab:results} shows cross validation results for all investigated models.
Precision, recall and $F_1$ derived from the SoftED metrics are listed.
To convey an impression of size, the number of parameters for each architecture is given.

\begin{table}[tb]
	\centering
	\caption{Results of cross validation}
	\label{tab:results}

	\begin{tabular}{lcccr}
		\toprule
		architecture & precision & recall & $F_1$ & \# parameters \\ \midrule
		CNN + dense & \textbf{0.93} & \textbf{0.93} & \textbf{0.93} & 513,317,544 \\
		CNNc1 & 0.70 & 0.71 & 0.70 & 1,317,057 \\
		CNNc2 & 0.81 & 0.81 & 0.81 & 6,284,353 \\
		CNNc3 & 0.89 & 0.88 & 0.88 & 10,476,545 \\ \midrule
		GRUc1 & 0.86 & 0.84 & 0.85 & 37,889 \\
		BGRUc1 & \textbf{0.93} & 0.91 & 0.92 & 100,353 \\
		BGRUc2 & \textbf{0.93} & 0.92 & \textbf{0.93} & 249,345 \\
		BGRUc3 & \textbf{0.93} & 0.92 & \textbf{0.93} & 990,209 \\
		\bottomrule
	\end{tabular}
\end{table}

\subsection{CNN results}\label{cnn-results}

The results for CNN-based models are shown in the upper part of Tab.~\ref{tab:results}.
Each model features similar precision and recall.
The $F_1$ score is the harmonic mean of recall and precision and is used for further model comparison.
The best CNN-based performance is achieved for the \textit{CNN + dense} model which provides an $F_1$ score of 0.93.
The \textit{CNNcX} series yields $F_1$ scores between 0.70 and 0.88 with the score increasing with depth and numbers of parameters of the model. 

\subsection{RNN results}\label{rnn-results}

RNN-based results are summarized in the lower part of Tab.~\ref{tab:results}.
As with the CNNs, precision and recall are similar for each model. The \textit{GRUc1} is the model with by far the fewest parameters among the tested models and still provides an $F_1$ score of 0.85. All \glspl{bgru} yield high $F_1$ scores at 0.92 to 0.93.
As before, greater depth goes along with a higher number of parameters but not necessarily with an better prediction score.

\subsection{CNN vs.~RNN}\label{cnn-vs.-rnn}

\begin{table}[tb]
	\centering
	\caption{Results of final evaluation}
	\label{tab:final_results}

	\begin{tabular}{lccc}
		\toprule
		architecture & precision & recall & $F_1$ \\ \midrule
		CNN + dense & 0.92 & \textbf{0.93} & 0.92 \\
		BGRUc2 & \textbf{0.94} & \textbf{0.93} & \textbf{0.93} \\
		\bottomrule
	\end{tabular}
\end{table}

The \textit{CNN + dense} model is the best among all CNNs and therefore used for comparison with RNN-based approaches.
Among the RNNs, \textit{BGRUc2} and \textit{BGRUc3} perform equally well while the former has fewer parameters.
Since less parameters are beneficial for practical applications (see Sec.~\ref{discussion}), the \textit{BGRUc2} model is used for the final comparison of \gls{rnn}- and \gls{cnn}-based approaches as described in Sec.~\ref{Experimental-setup-and-target-metrics}.

Table~\ref{tab:final_results} shows the results of the final evaluation using the holdout dataset.
Again, only small differences between precision and recall can be observed.
The \textit{CNN + dense} architecture yields an $F_1$ score of 0.92 while \textit{BGRUc2} features an $F_1$ score of 0.93.

\begin{figure}[tb]
	\centering
	\includegraphics[width=0.9\linewidth]{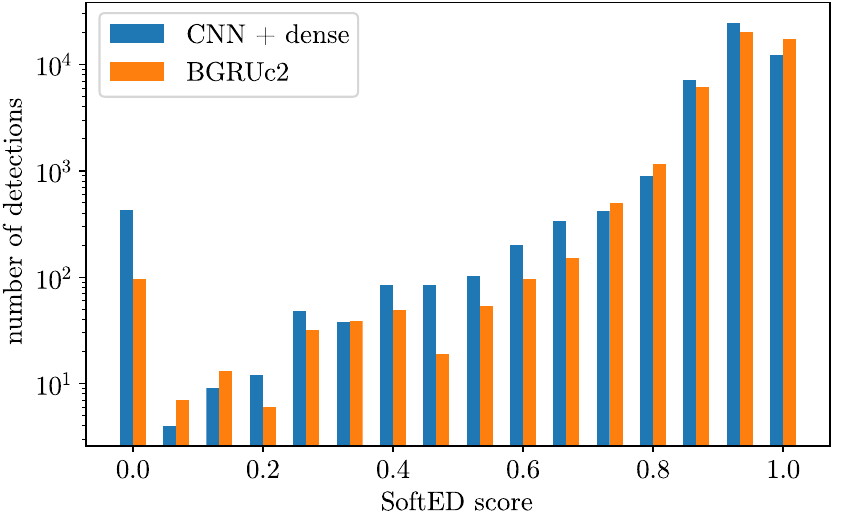}
	\caption{Comparison of the distribution of SoftED scores in the final evaluation. The y-axis is scaled logarithmically.}
	\label{fig:histogram_final}
\end{figure}

Fig.~\ref{fig:histogram_final} shows the distributions of SoftED scores for the final evaluation.
While the \textit{CNN + dense} model gives more predictions scored 0 and fewer detections scored 1, it also yields more detections scored 0.87 or 0.93 than the \textit{BGRUc2} model.
In fact, no clear tendency for certain strengths or weaknesses among the two final models is observed.

\section{Discussion}\label{discussion}

In this work we employed CNN- and RNN-based models to detect onsets and endings of paddle strokes in a force sensor signal. These events cannot be defined in a straight analytical manner but are usually derived iteratively by human experts.
Our goal was to assess whether these models are generally suitable to detect these events and if the use of a certain architecture is beneficial over the other for this task. We conducted several experiments and measured the model performance by introducing modifications to the recently published SoftED metric~\cite{Salles2023}. Our modifications improve the metric for use with signals derived using the sliding window method.

Our experiments included four CNN-based models. Three of these consisted of convolutional layers only (\textit{CNNcX}), while the \textit{CNN + dense} model used a fully connected layer before the output. This not only resulted in a significant increase in performance compared to other CNNs, but made it to the best performing model in our work. Hence, the use of the dense layer with CNN-based models seems very beneficial for this task. The resulting number of parameters, however, is huge, especially when compared to any other model. This may reduce the suitability of \textit{CNN + dense} for practical applications in, e.\,g., real-time environments.

The experiments conducted using RNNs included one unidirectional (GRU) and three bidirectional models (BGRUs). The performance of the latter is similar to the best CNN model. Notably, all these models require 2+ order of magnitudes fewer parameters than the best CNN, and even only a fraction of those as used by the non-dense layer CNNs while the BGRUs detection performances are much better. It also turned out that increasing the depth of BGRU models only results in a minor performance increase. Overall, RNN-based models seem beneficial compared to CNNs due to their similar performance with much fewer parameters.

The best performing model gave an F1 of 0.93 and there may be room for further improvement (see below). However, due to the underlying nature of the problem, i.\,e. the prediction of human expert labels, a much higher performance might not be expected for various reasons. First, the inter-rater variability leads to noise in human expert labels. Second, there was evidence in our results that some labels were incorrect or even missing. This is due to the current practice of considering the events holistically rather than individually, which is not applicable. Although the impact of these problems is certainly small, they do place an upper limit on the detection 
accuracy.

A further increase in performance using CNNs might be achieved by optimizing the receptive field of the models while ensuring a gradual increase in the number of parameters. For example, increasing the number of layers and adjusting dilation within layers might improve results for convolutional-only models and could reduce the number of parameters in a final dense layer, if used. However, this kind of hyperparameter selection was not investigated in this work. In fact, there was no sophisticated hyperparameter selection at all. Therefore, small differences in model performance, such as those reported in Sec.~\ref{results}, may be due to suboptimal parameter choices rather than to the model itself. Finally, further improvements might be achieved by varying the decision threshold for event extraction from the predicted outputs.

Our results show that human expert labels in force sensor signals can automatically be detected using CNNs and RNNs. Based on our method, the detection can be performed on temporal segments with only one or multiple paddle strokes. This is important for targeting real-time applications in future research, i.\,e. analysing paddle stroke-based parameters even before a trial is complete, e.g. for live feedback. 

Finally, the methods investigated in this work are certainly not limited to the analysis of force sensor signals in canoe and kayak sprint but could be applied to any sensor signals and, more importantly, to other fields of application both inside and outside the field of sport. Our study shows that events which are mainly defined by human experts rather than by a sophisticated analytical definition - a scenario often found in the field of sports science - can be automatically detected by exploiting the predictive power of machine learning. This is particularly interesting when models can be trained and evaluated using historical data without the need for new measurements, which also is often the case in sports.

\bibliographystyle{IEEEtran}
\bibliography{bibliography.bib}
\vspace{12pt}
\end{document}